\PassOptionsToPackage{table}{xcolor} % 加上 table 选项以支持表格颜色
\documentclass[sigconf, screen]{acmart}

\AtBeginDocument{%
  }

\usepackage{multirow}

\begin{document}

\title{MCA-LLaVA: Manhattan Causal Attention for Reducing Hallucination in Large Vision-Language Models}

\author{Qiyan Zhao}
\authornote{represents the co-first author. \ddag represents project leader.}
% \authornote{represents the co-first author. This work was done during the Qiyan Zhao's internship at Fujian Key Laboratory of Pattern Recognition and Image Understanding, XMUT.}
\affiliation{%
  \institution{FKLPRIU, Xiamen University of Technology, China}
  % \institution{Fujian Key Laboratory of Pattern Recognition and Image Understanding}
  \country{}}
\email{qiyanzhao618@gmail.com}

\author{Xiaofeng Zhang \dag \ddag}
\authornotemark[1]
\affiliation{%
  \institution{Shanghai Jiao Tong University, China}
  \country{}}
\email{SemiZxf@163.com}

\author{Yiheng Li}
\affiliation{
  \institution{Nanyang Technological University, Singapore}
  \country{}}
\email{yiheng003@e.ntu.edu.sg}

\author{Yun Xing}
\affiliation{
  \institution{Nanyang Technological University, Singapore}
  \country{}}
\email{xing0047@e.ntu.edu.sg}

\author{Xiaosong Yuan}
\affiliation{
  \institution{Jilin University, China}
  \country{}}
\email{yuanxs19@mails.jlu.edu.cn}

\author{Feilong Tang}
\affiliation{%
  \institution{Monash University, Australia}
  % \institution{Fujian Key Laboratory of Pattern Recognition and Image Understanding}
  \country{}}
\email{feilong.tang@monash.edu}

\author{Sinan Fan}
\affiliation{%
  \institution{Zhejiang University, China}
  \country{}}
\email{222064@zju.edu.cn}

\author{Xuhang Chen}
\affiliation{
\institution{Huizhou University, China}
\country{}
}
\email{xuhangc@hzu.edu.cn}

\author{Xu-Yao Zhang}
\affiliation{%
  \institution{Chinese Academy of Sciences, China}
  % \institution{Fujian Key Laboratory of Pattern Recognition and Image Understanding}
  \country{}}
\email{xyz@nlpr.ia.ac.cn}

\author{Da-Han Wang\dag}
\thanks{\dag represents the corresponding author. This work is supported by the Major Science and Technology Plan Project on the Future Industry Fields of Xiamen City (No. 3502Z20241027), the Unveiling and Leading Projects of Xiamen (No. 3502Z20241011) , and the Open Project of the State Key Laboratory of Multimodal Artificial Intelligence Systems (MAIS2024101).}
\affiliation{%
  \institution{FKLPRIU, Xiamen University of Technology, China}
  % \institution{Fujian Key Laboratory of Pattern Recognition and Image Understanding}
  \country{}}
\email{wangdh@xmut.edu.cn}

\begin{abstract}
Hallucinations pose a significant challenge in Large Vision Language Models (LVLMs), with misalignment between multimodal features identified as a key contributing factor. This paper reveals the negative impact of the long-term decay in Rotary Position Encoding (RoPE), used for positional modeling in LVLMs, on multimodal alignment. Concretely, under long-term decay, instruction tokens exhibit uneven perception of image tokens located at different positions within the two-dimensional space: prioritizing image tokens from the bottom-right region since in the one-dimensional sequence, these tokens are positionally closer to the instruction tokens. This biased perception leads to insufficient image-instruction interaction and suboptimal multimodal alignment. We refer to this phenomenon as “image alignment bias.” To enhance instruction's perception of image tokens at different spatial locations, we propose MCA-LLaVA, based on Manhattan distance, which extends the long-term decay to a two-dimensional, multi-directional spatial decay. MCA-LLaVA integrates the one-dimensional sequence order and two-dimensional spatial position of image tokens for positional modeling, mitigating hallucinations by alleviating image alignment bias. Experimental results of MCA-LLaVA across various hallucination and general benchmarks demonstrate its effectiveness and generality. The code can be accessed in \url{https://github.com/ErikZ719/MCA-LLaVA}.
% 并阐明了根据一维序列中tokens的相对位置计算的long-term decay不适应二维图像中信息分布的空间特点

% 根本原因是long-term decay按照一维，单方向的衰减趋势
\end{abstract}





\maketitle

\begin{figure}[htb]
% \vspace{-10pt}
\centering
\centerline{\includegraphics[width=8.5cm]{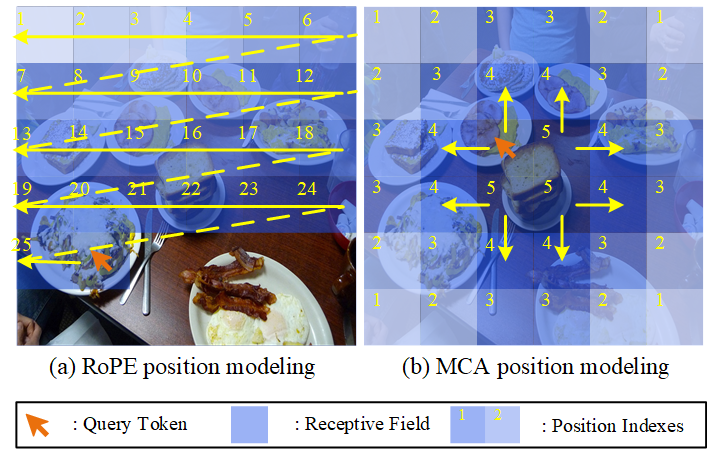}}
\caption{Schematic of long-term decay in different positional encoding mechanisms. Yellow arrows indicate the direction of decay. In the causal attention mechanism, the direction of decay always goes from tokens with larger position indices to tokens with smaller position indices. (a) denotes RoPE one-dimensional unidirectional long-term decay. (b) denotes MCA spatial multi-directional long-term decay. Darker colors represent smaller decay, while lighter colors represent larger decay. The number of encoded image tokens is set to 36 for the demonstration.}
\label{motivation}
\vspace{-13pt}
\end{figure}

\section{Introduction}

Large Vision-Language Models (LVLMs) \cite{llava, llava1.5, bai2023qwen, GPT4V, internvl,Qwen2-VL, gpt4roi, mplug-owl2} have demonstrated impressive multimodal understanding across various domains, such as document comprehension \cite{gqa, docvqa} and complex visual reasoning \cite{mme}. However, the reliability of LVLMs is compromised by hallucinations \cite{chair, pope}, a phenomenon in which models generate counterfactual responses that do not align with the information from the question image.

Recent studies \cite{align, pai, survey2} have identified the misalignment between visual and textual inputs as a key contributor to hallucinations. In particular, widely used Large Vision-Language Models (LVLMs) typically project encoded visual features into the textual embedding space of Large Language Models (LLMs) \cite{llama}. However, the inherent distribution gap between visual and textual tokens poses significant challenges for cross-modal interaction and feature alignment. To address this issue and reduce hallucinations, several approaches have focused on enhancing modality alignment through improved alignment training, such as augmenting fine-grained data for instruction fine-tuning \cite{Liu2023MitigatingHI}, or applying reinforcement learning guided by human feedback \cite{sun2023aligning}. More recently, a series of contrastive decoding strategies \cite{vcd, ibd, icd} has been introduced to mitigate hallucinations during inference without additional training. Although these methods offer partial relief, the internal mechanisms within the models that lead to modality misalignment and hallucinations remain insufficiently understood.

Recent research focusing on information flow has provided insight into relationship between hallucinations and LVLMs internal mechanisms. Label Words \cite{label-words} first combines the attention value and gradient to identify attention sink observing that the information flow always converges to the prompt token eventually. OPERA\cite{opera} attributed object hallucination to some prompt tokens that receive consistently high attention at the decoding stage. EAH\cite{eah} found that input image tokens with high-density information flow in some attention heads can help mitigate hallucination. Leveraging information flow, these studies clarify the relationship between prompt tokens, image tokens, and hallucinations. However, the information flow between tokens of different modalities remains to be explored. A deeper exploration of multimodal information flow may help to reflect the multimodal alignment process in LVLMs. 

Unlike previous work, this paper delves into RoPE \cite{ROPE}, a key component for position modeling in LVLMs, examining its negative impact on multimodal  alignment and its relationship with hallucinations.

\noindent
\textit{\textbf{Q1: Is long-term decay of RoPE suitable for multimodal alignment?}}

Under RoPE long-term decay, varying levels of attention are assigned to image tokens based on their relative distances from instruction tokens, with those farther away receiving less attention. To investigate information interaction under long-term decay, we visualize the image-to-instruction information flow, as shown in Figure \ref{informationflow}.a. We found that only the tokens in the lower-right region of the image exhibited dense information flow due to their proximity to the instruction tokens in the one-dimensional sequence, while a large number of image tokens, distant from the instruction tokens, showed very sparse information flow. This imbalanced distribution of information flow further reveals the model's uneven perception: only image tokens in a limited region interact sufficiently with the instruction tokens, which hinders multimodal alignment and leads to hallucinations. We refer to this phenomenon as "image alignment bias" and provide a detailed explanation in Section 3.

\noindent
\textit{\textbf{Q2: What limits the long-term decay of RoPE in multimodal alignment?}}

CCA-LLaVA ~\cite{cca-llava} is the first to find that LVLMs may be more prone to hallucinations because of long-term decay. To this end, it heuristicly redirects instruction tokens to focus more on the image center region by reassigning the image tokens' position indices in the form of concentric squares. Distinct from CCA, we observe that long-term decay disregards the two-dimensional position of the image, considering only the image tokens' position in the one-dimensional sequence when calculating relative distances. As shown in Figure \ref{motivation}.a, this distance-dependent decay aligns with the order distribution of information in a 1D text sequence but overlooks the spatial distribution of information in a 2D image. Therefore, our goal is to extend the long-term decay to the two-dimensional spatial domain, calculating the relative positional distances of image tokens based on spatial locality. This enhances the model's perception of image tokens at different spatial locations.

To this end, we propose Manhattan Causal Attention (MCA) to mitigate hallucination in LVLMs caused by RoPE long-term decay. MCA consists of three key designs: 1. The RoPE one-dimensional long-term decay is evolved into the two-dimensional, multi-directional spatial decay by calculating the Manhattan distance between tokens; 2. We reassign the two-dimensional position coordinates of image tokens to align with the Manhattan distance computation. After that we replace the RoPE raster-scan position indices with new position indices, which are computed based on position coordinates; 3. Modeling image position dependency by Manhattan causal mask module, which preserves 2D spatial localization properties. 

We evaluated the MCA extensively: compared to the baseline, MCA improves F1 scores by +6.7\% and Accuracy by +6.7\% on POPE, and reduces sentence-level hallucination by 9\% and instance-level hallucination by 2.9\% on CHAIR. Additionally, our approach enhances overall image perception and information interaction, showing promising performance on several general tasks such as MME and SQA. Experimental results across multiple hallucination benchmarks and a range of LVLMs show the consistent improvements brought by MCA. Our contribution consists of three parts:

% Besides using MCA to train LVLMs models, we found that directly reorganizing the image tokens' positions in a train-free way can also alleviate the object hallucination of LVLMs. We evaluated the MCA extensively: in the train-free situation, compared to baseline, the MCA improves F1 scores by +5.9\% and Accuracy +5.8\% on POPE, and reduces sentence-level hallucination by -5.4\% and instance-level by -1.3\% on CHAIR. Our trained model with MCA shows better performance on several hallucination benchmarks. Additionally, our approach enhances overall image perception and information interaction, showing promising performance on multiple general cross-modal tasks. Our contribution consists of three parts: 

% The key insight of MCA is to clarify that temporal decay is caused by the results of the RoPE relative position calculation, and that improvements should be investigated from that calculation mechanism. 

% RoPE \cite{ROPE} models position dependencies based on the relative distance between tokens, which introduces temporal decay to causal attention $Attn$ as follows:
% \begin{equation}
% \resizebox{\columnwidth}{!}{$
% Attn(Q_n, K_m) = \left(Softmax \left(\frac{Q_n^T \cdot D^{n-m}\cdot K_m}{\sqrt{d}} \right)\right)
% $ }
% \end{equation}
% Here, $D_{n-m}$ represents the position encoding, which depends on the relative distance $n-m$ between the query token $Q_n$ and the key token $K_m$.

% The temporal decay refers to $Attn(Q_n, K_m)$随着相对距离$n-m$减少

\begin{itemize}
    
\item
This paper delves into the relationship between RoPE and hallucinations. We reveal the image alignment bias, arising from RoPE long-term decay, leads to inferior multimodal alignment and hallucinations.

\item  We propose Manhattan Causal Attention, which models image position dependency by Manhattan relative position distance. MCA extends the long-term decay to the 2-D multi-directional spatial decay, which mitigates hallucinations caused by image alignment bias.

\item Experiments on both hallucination and general benchmarks demonstrate the promising performance of our design.

\end{itemize}

% The ``\verb|acmart|'' document class can be used to prepare articles
% for any ACM publication --- conference or journal, and for any stage
% of publication, from review to final ``camera-ready'' copy, to the
% author's own version, with {\itshape very} few changes to the source.

\section{Related Work}

\subsection{Hallucination in LVLMs}

Hallucination in Large Vision-Language Models (LVLMs) refers to the phenomenon in which the model's textual output contradicts the visual input, such as generating descriptions that include objects or attributes not present from the image \cite{pope, chair, ccm, mm-vet}. Most existing LVLMs project encoded visual features into the input space of the language model; however, a significant modality gap between textual and visual tokens often results in cross-modal misalignment, leading to hallucinations during generation \cite{survey2, align, li2023evaluating, zhou2023analyzing, gunjal2024detecting}. Several studies have sought to improve cross-modal alignment interfaces to reduce hallucination risks \cite{llava1.5, internvl, He2024IncorporatingVE, Shang2024FromPT}, while others have employed contrastive learning strategies to enhance alignment between visual and textual representations \cite{Jiang2023HallucinationAC, Sarkar2024MitigatingOH, Li2025TheHL}. Additional approaches have leveraged diverse, fine-grained fine-tuning datasets \cite{HalluciDoctor, ccm, Liu2023MitigatingHI, Chen2025PerturboLLaVARM, Zhang2024ReflectiveIT, Wang2023MitigatingFH} or human feedback alignment \cite{sun2023aligning, yu2023rlhf}, though these typically incur high annotation costs. Recently, a range of decoding strategies has been proposed to mitigate hallucinations by intervening in the model’s reasoning process without requiring further training \cite{ibd, vcd, icd, hio, code, VACoDe, Gong2024DAMRODI, Wang2024VaLiDMT}; however, these methods do not improve intrinsic modal alignment and often reduce inference efficiency. Preference optimization techniques, which train models based on comparisons between positive and negative samples, have also been explored \cite{Pi2024StrengtheningML, Fu2024MitigatingHI, Xie2024VDPOMH, Wang2024mDPOCP, Cui2024FineGrainedVP, Zhou2024CalibratedSV, Wang2024CREAMCR, Xiong2024LLaVACriticLT}, though these are often prone to overfitting on specific datasets. Recent research has further revealed that LVLMs exhibit attention bias between image and text modalities \cite{ClearSight, pai, Zhu2024UnravelingCK, cca-llava}, where insufficient visual perception contributes to hallucination. Some methods address this issue by correcting attention distribution bias in a training-free manner \cite{ClearSight, eah, agla,tang1,tang2,tang3}, while others adopt fine-tuning strategies to enhance multimodal alignment \cite{cca-llava, achallu}. These approaches have demonstrated improved attention to visual input. In contrast to the above methods, this paper investigates the impact of RoPE’s long-term decay on cross-modal alignment, from the perspective of positional modeling mechanisms in LVLMs.

\subsection{Information Flow in LVLMs}

With the rapid development of LVLMs\cite{li2024llava, bai2023qwen, GPT4V, Qwen2-VL, radford2021learning,zhang2023video}, more and more works try to find inspiration for model optimization by analyzing the internal mechanisms of the models. Among these, information flow \cite{label-words, fastv,zhang-aaai,zhang-nn} provides an intuitive method to understand the internal mechanisms of LVLMs black-box models. Label Words \cite{label-words}, and ACT \cite{attention-sink} are early works that explore the mechanism of LLMs\cite{minigpt4, devlin2018bert, llama} through observing information flow patterns. By calculating saliency scores, it is possible to visualize the information flow. 

Building on this, OPERA \cite{opera} and DOPRA \cite{DOPRA} introduced information flow to reveal the relationship between token attention value and hallucinations. They found that during the decoding of LVLMs, some special tokens (e.g., “-”, “? “) receive consistently high attentional values, which leads to hallucinations. To this end, they propose different penalty constraints to alleviate the over-reliance on these tokens. LLaVA-CAM \cite{zhang-nips}combines Grad-CAM and attention map to propose a dynamic analysis of information flow, which reveals the fine-grained effect of token in the LVLMs prediction. EAH \cite{eah} analyzed the information flow of image tokens across each layer and head of the LVLMs. It proposes a train-free method to enhance the information flow distribution of image tokens in specific layers to improve the image perception of the model.
% Building on this, OPERA \cite{opera} and DOPRA \cite{DOPRA} introduced information flow to reveal the relationship between token attention value and hallucinations. They found that during the decoding of LVLMs, some special tokens (e.g., “-”, “? “) receive consistently high attentional values, which leads to LVLMs hallucinations. To this end, OPERA and DOPRA propose different penalty constraints to alleviate the over-reliance on these tokens, effectively reducing hallucinations. LLaVA-CAM \cite{zhang-nips}combines Grad-CAM and attention map to propose a dynamic analysis of information flow, which reveals the fine-grained effect of token in the LVLMs prediction. EAH \cite{eah} analyzed the information flow of image tokens across each layer and head of the LVLMs. It proposes a train-free method to enhance the information flow distribution of image tokens in specific layers to improve the image perception of the model.

% Some studies have proposed a series of valuable interpretable methods and hallucination mitigation methods by analyzing the information flow of LVLMs.
% \vspace{-5pt}
\subsection{Position Encoding in LVLMs}

Position encoding was first proposed for Transformer\cite{transformer} sequence tokens position dependency modeling. To enable LVLMs to understand the order information of tokens, different methods have been proposed to encode position information into representations. Commonly used position encoding includes absolute position encoding\cite{transformer}, learnable position encoding\cite{VIT}, and relative position encoding\cite{RelaRPE}. QwenVL2\cite{Qwen2-VL} proposes Multimodal Rotary Position Embedding to incorporate temporal information into the position representation. Among them, RoPE\cite{ROPE} encodes positional representation for linear attention using rotation matrices and is widely used in LVLMs. This paper provides an in-depth analysis of the relationship between RoPE and hallucination phenomenon and reveals the limitations of the relative position calculation.

\subsection{Image Alignment Bias}

In this section, we first analyze the phenomenon of image alignment bias triggered by the RoPE long-term decay. By further analyzing the image-to-instruction information flow, we reveal the negative effects of image alignment bias on multimodal alignment and hallucination. Finally, we clarify the causes and limitations of the image alignment bias from the perspective of the RoPE positional encoding mechanism.

% \begin{figure}[htb]
% \centering
% \centerline{\includegraphics[width=7cm]{pilot experiment.png}}
% \caption{pilot experiment.}
% \label{fig:res}
% %
% % \vspace{-15pt}
% \end{figure}

\noindent
\textbf{Long-term decay causes image alignment bias:}
Under the long-term decay induced by RoPE, target tokens positioned farther from an instruction token experience greater decay in their attention scores \cite{ROPE}. This decay aligns with the typical distribution of information in language modeling, wherein text is represented as a one-dimensional sequence and tokens closer in relative position to the query token generally carry more consistent and semantically relevant information. 

\begin{figure*}[htb]
% \vspace{-10pt}
\centering
\centerline{\includegraphics[width=14cm]{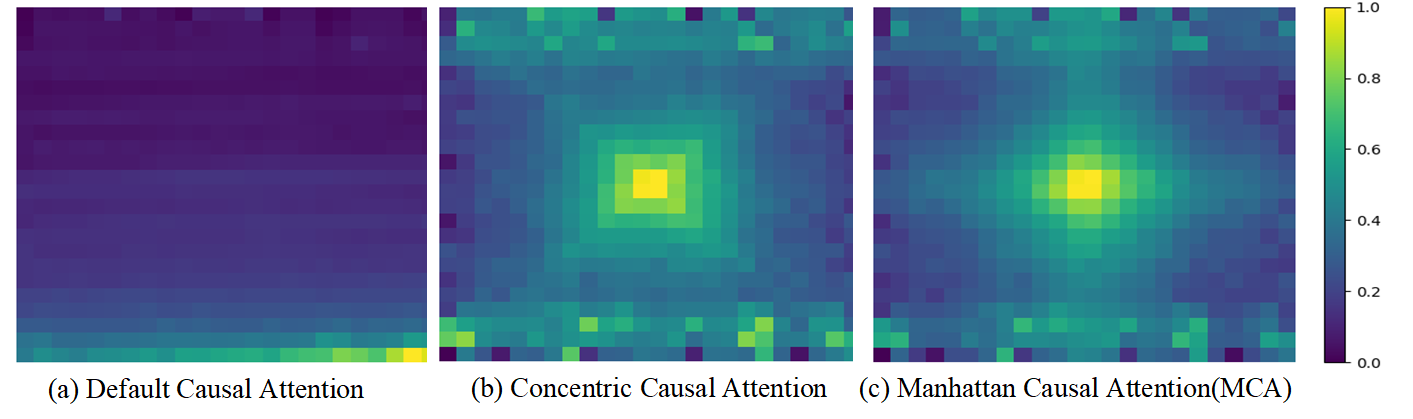}}
\caption{(a), (b), and (c) show the information flow of image-to-instruction in LLaVA1.5\cite{llava1.5}, CCA\cite{cca-llava}, and MCA, respectively. We aggregate the information flow of each image token in the input tokens to the instruction tokens, and the aggregation results are arranged according to the corresponding positions of the image tokens in the 2-D space. The reported statistics are averaged over the 3K adversarial subset used for evaluation in the POPE \cite{pope}.}
\label{informationflow}
\vspace{-5pt}
\end{figure*}

In LVLMs, image tokens are flattened into a 1-D sequence using a raster-scan order (top-to-bottom, left-to-right) and concatenated with instruction tokens to form the input sequence. Due to RoPE's decay effect, attention toward image tokens positioned farther from the instruction tokens decays progressively. This results in a fixed multimodal alignment pattern in which instruction tokens predominantly attend to image tokens located later in the raster-scan order, while tokens earlier in the sequence receive limited attention, as illustrated in Figure \ref{motivation}.a. We term this phenomenon as image alignment bias, which is a systemic bias introduced into the visual feature attention by the internal mechanisms of LVLMs, rather than an attention pattern learned based on training.

\noindent
\textbf{Information flow for image alignment bias:}
We visualize the image-to-instruction information flow to examine the impact of image alignment bias on multimodal alignment. As shown in Figure \ref{informationflow}.a, we find that image tokens in the lower-right region, which are closer to the instruction token, exhibit dense information flow, while the majority of image tokens in other regions exhibit sparse information flow. This suggests that many image tokens fail to interact sufficiently with instruction token, which leads to multimodal misalignment and hallucinations.

This extreme information flow distribution hinders the model's perception of overall image information. CCA was the first to identify a similar phenomenon, confirming that LVLMs are more likely to generate hallucinations when relevant visual cues are positioned far from instruction tokens within the multimodal input sequence. To mitigate this issue, a heuristic reallocation of visual attention was proposed. However, this approach lacks interpretability with respect to the internal mechanisms of LVLMs. In this work, the underlying causes of unbalanced information flow and image alignment bias are analyzed through the lens of RoPE long-term decay.

\noindent
\textbf{Ignoring the spatial properties of the image in relative position calculations during long-term decay:} Given a multimodal input sequence of LVLMs, $Q_i$ is the instruction query token at position i and $K_j$ is the image key token at position j. To model the relative position dependency among them, RoPE multiplies the $Q_i$ and $K_j$ with the rotation matrix via $R^d_{\theta,i} \cdot Q_i$ and $R^d_{\theta,j} \cdot K_j$. The rotation matrix $R^d_{\theta,m}$ is shown in Eq. 1, where\begin{equation}
\resizebox{\columnwidth}{!}{
  $\displaystyle R^d_{\theta,m} = \begin{pmatrix}
  \cos(m\theta_1) & -\sin(m\theta_1) & 0 & 0 & \dots & 0 & 0\\
  \sin(m\theta_1) & \cos(m\theta_1) & 0 & 0 & \dots & 0 & 0\\
  0 & 0 & \cos(m\theta_2) & -\sin(m\theta_2) & \dots & 0 & 0\\
  0 & 0 & \sin(m\theta_2) & \cos(m\theta_2) & \dots & 0 & 0\\
  \vdots & \vdots & \vdots & \vdots & \ddots & \vdots & \vdots\\
  0 & 0 & 0 & 0 & \dots & \cos(m\theta_{d/2}) & -\sin(m\theta_{d/2}) \\
  0 & 0 & 0 & 0 & \dots & \sin(m\theta_{d/2}) & \cos(m\theta_{d/2})
  \end{pmatrix}$ 
  }
\end{equation}$\left\{ \theta_i = 10000^{-2(i-1)/d} \right\}$, $\quad i \in ( 1, 2, \dots, d/2)$ denotes the predefined sinusoidal function values, $d$ denotes the embedding dimension, and $m$ denotes the position index. The attention value $Attn$ between $Q_i$ and $K_j$ is calculated as follows:
\begin{equation}
\resizebox{\columnwidth}{!}{
$ Attn_{i,j} = \text{softmax}\left( \frac{Q_i^T \cdot \left( R^d_{\theta,i} \right)^T \cdot R^d_{\theta,j} \cdot K_j}{\sqrt{d}} \right) = \text{softmax}\left( \frac{Q_i^T \cdot R^d_{\theta,(j-i)} \cdot K_j}{\sqrt{d}} \right)
$ }
\end{equation} The $Attn$ reflects the degree of long-term decay: As the relative distance $j-i$ between image and instruction tokens increases, the attention $Attn$ gradually decreases. In calculating relative distances, image tokens are assigned position indices based on their order in a 1-D sequence after flattening, ignoring their 2-D spatial positions. As a result, the long-term decay induces image alignment bias.

\begin{figure*}[t]
\centering
\centerline{\includegraphics[width=18cm]{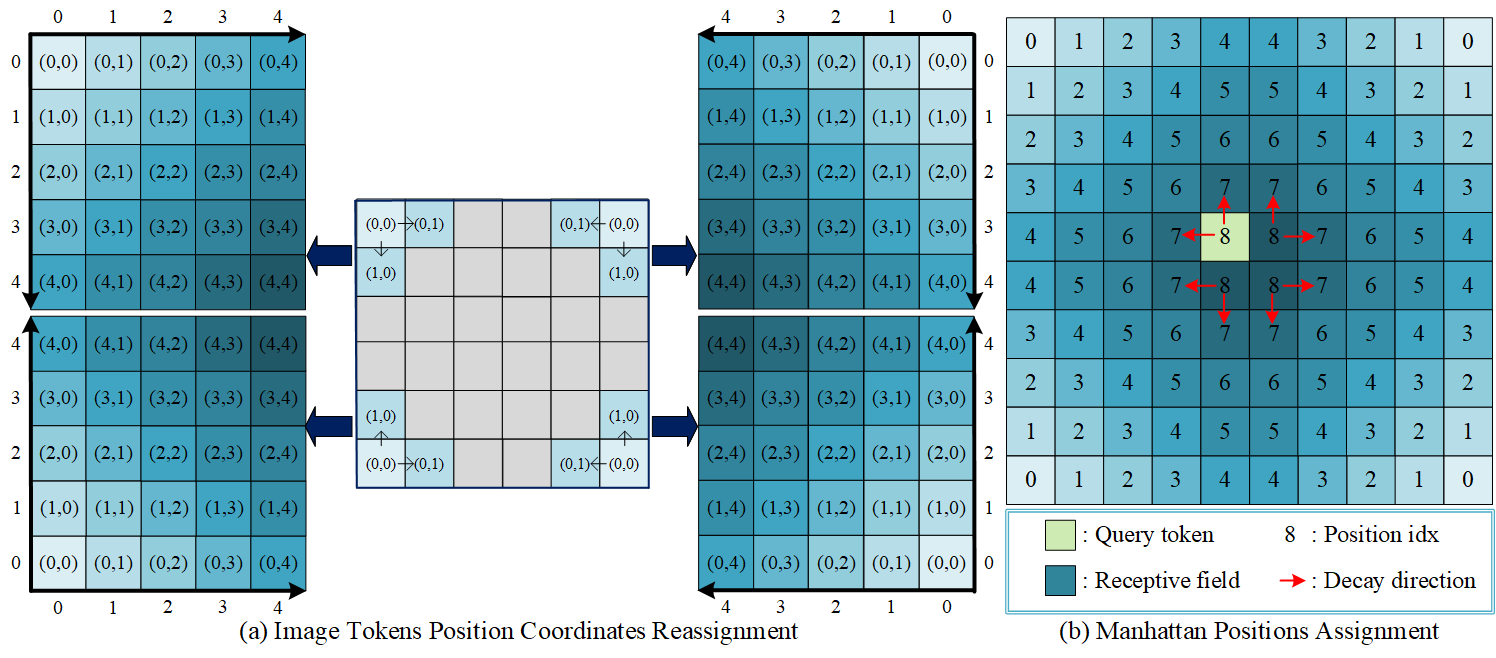}}
\caption{Illustration of image token position coordinate reassignment. The total number of image tokens is denoted as $V$; for demonstration purposes, $V=100$, while the default setting in LLaVA-1.5 uses $V=576$. The image tokens are mirrored into four partitions, with the four vertex positions designated as origins. Subsequently, two-dimensional positional coordinates are assigned sequentially to the remaining visual tokens based on the direction of the coordinate axes shown in (a). (b) shows the new position index computed from the coordinates.}
\label{coordinates}
%
% \vspace{-15pt}
\end{figure*}

\section{Manhattan Causal Attention}

To mitigate the object hallucination caused by the ROPE image alignment bias, we propose the Manhattan Causal Attention (MCA). MCA consists of three parts: 1. Relative position distance computation evolves from one-dimensional to two-dimensional Manhattan distances, preserving the spatial nature of the image; 2. Assigning position coordinates to each image token and merging the position coordinates as new position indexes; and 3. Modify the default causal attention masking to Manhattan causal masking.

\subsection{Manhattan Relative Position Distance}

As shown in Eq. 3, RoPE models the relative \begin{equation}
D_{RoPE} \{Q_i,K_j\} = \gamma(j) - \gamma(i)
% D_{ROPE} \{Q_i,K_j\} = j - i
\end{equation} distance between $Q_i$ and $K_j$ as $\gamma(j) - \gamma(i) $. $\gamma$ denotes the position index under raster scanning. This approach limits the relative positional distances of image tokens to the one-dimensional level and loses the spatial locality of the two-dimensional image. 

To address this limitation, we extend the computation of the relative positional distance of image tokens to two-dimensional levels. Naturally, the image tokens at position $m$ can correspond to a coordinate in two-dimensional space $(x_m,y_m)$. Therefore we expand the one-dimensional relative positional distance (Eq. 3) to the two-dimensional Manhattan relative positional distance (Eq. 4) between the image tokens coordinates.

\begin{equation}
D_{Manhattan} \{Q_i,K_j\} = (x_{j} - x_{i})+(y_{j} - y_{i})
\end{equation}

\noindent
\textbf{From Position indexes to Position Coordinates:}
Under raster-scan, image tokens are scanned row by row starting from the top left and assigned position indexes with increments of 1 as follows: 
\begin{equation}
\resizebox{\columnwidth}{!}{
  $\left[ 
\begin{array}{ccccccccc}
0 & 1 &  & &\cdots & & &\sqrt{v}-2 & \sqrt{v}-1 \\
\sqrt{v} & \sqrt{v}+1  & & & \cdots & && 2\sqrt{v}-2 &2\sqrt{v}-1\\
& & \ddots&  & & && & \\
\vdots & \vdots & & \frac{v}{2}-\frac{\sqrt{v}}{2}-1& &\frac{v}{2}-\frac{\sqrt{v}}{2} &   & \vdots & \vdots \\
\vdots & \vdots &  & \frac{v}{2}+\frac{\sqrt{v}}{2}-1 && \frac{v}{2}+\frac{\sqrt{v}}{2}&   & \vdots & \vdots \\
&  & & & & & \ddots& & \\
v-2\sqrt{v} & v-2\sqrt{v}+1 &  & & \cdots& && v-\sqrt{v}-2 & v-\sqrt{v}-1\\
v-\sqrt{v} & v-\sqrt{v}+1 &  & &\cdots &&& v-2 & v-1 
\end{array}
\right]$ 
  }
\end{equation}
where $v$ denotes the number of image tokens. Since adding column coordinates directly to the position index cannot compute $D_{Manhattan}$ correctly, we reassign coordinates to the image tokens. 

% \begin{equation}
% \resizebox{\columnwidth}{!}{
%   $\left[ 
% \begin{array}{ccccccccc}
% 0 & 0 &  & &\cdots & & &0 & 0 \\
% 0 & 1  & & & \cdots & && 1 &0\\
% & & \ddots&  & & && & \\
% \vdots & \vdots & & \frac{\sqrt{v}}{2}-1& &\frac{\sqrt{v}}{2}-1 &   & \vdots & \vdots \\
% \vdots & \vdots &  & \frac{\sqrt{v}}{2}-1 && \frac{\sqrt{v}}{2}-1&   & \vdots & \vdots \\
% &  & & & & & \ddots& & \\
% 0 & 1 &  & & \cdots& && 1 & 0\\
% 0 & 0 &  & &\cdots &&& 0 & 0 
% \end{array}
% \right]$ 
%   }
% \end{equation}

Specifically, the tokens at the four vertices of the image are set as the origin points with coordinate $(0,0)$. As shown in Figure \ref{coordinates}.a, tokens adjacent to the origin points are treated as next tokens with an increment of 1. The positive direction of the horizontal and vertical coordinate axes is defined according to the incremental direction. The final image tokens' position coordinates are as follows:
\begin{equation}
\resizebox{\columnwidth}{!}{
  $\left[ 
\begin{array}{cccccccc}
(0,0) & (0,1) & \cdots & (0,\frac{\sqrt{v}}{2}-1) &(0,\frac{\sqrt{v}}{2}-1) &\cdots &(0,1) & (0,0) \\
(1,0) & (1,1)  & \cdots& (1,\frac{\sqrt{v}}{2}-1) &(1,\frac{\sqrt{v}}{2}-1) &\cdots& (1,1) & (1,0)\\
\vdots& \vdots& \ddots& \vdots &\vdots & & \vdots& \vdots\\
(\frac{\sqrt{v}}{2}-1,0) & (\frac{\sqrt{v}}{2}-1,1) & \cdots& (\frac{\sqrt{v}}{2}-1,\frac{\sqrt{v}}{2}-1) &(\frac{\sqrt{v}}{2}-1,\frac{\sqrt{v}}{2}-1) &  \cdots & (\frac{\sqrt{v}}{2}-1,1) & (\frac{\sqrt{v}}{2}-1,0) \\
(\frac{\sqrt{v}}{2}-1,0) & (\frac{\sqrt{v}}{2}-1,1) & \cdots& (\frac{\sqrt{v}}{2}-1,\frac{\sqrt{v}}{2}-1) &(\frac{\sqrt{v}}{2}-1,\frac{\sqrt{v}}{2}-1) &  \cdots & (\frac{\sqrt{v}}{2}-1,1) & (\frac{\sqrt{v}}{2}-1,0) \\
\vdots& \vdots & & \vdots& \vdots & \ddots&\vdots & \vdots\\
(1,0) & (1,1)  & \cdots& (1,\frac{\sqrt{v}}{2}-1) &(1,\frac{\sqrt{v}}{2}-1) &\cdots& (1,1) & (1,0)\\
(0,0) & (0,1) & \cdots & (0,\frac{\sqrt{v}}{2}-1) &(0,\frac{\sqrt{v}}{2}-1) &\cdots &(0,1) & (0,0) 
\end{array}
\right]$ 
  }
\end{equation}

The $V\times V$ image tokens are divided into four parts of the mirror image according to the four origin points, and each part consists of $\frac{\sqrt{v}}{2}\times\frac{\sqrt{v}}{2}$ tokens. In each part, the token's position coordinates are linearly incremented in the positive direction along the two-dimensional coordinate axis of the origin point, which preserves the two-dimensional localized spatial characteristics of images.

\noindent
% \textcolor{red}{\textbf{Manhattan Position Assignment:}}
\subsection{Manhattan Positions Assignment}
We note that by replacing the positional index of the raster-scan with the sum of the coordinate values of the tokens, the Manhattan relative positional distance can be formally aligned with Eq. 3. 
\begin{equation}
\begin{cases}
\mu(m) = x_{m} + y_{m}  \\
D_{Manhattan} \{Q_i,K_j\} = \mu(j) - \mu(i)
\end{cases}
\end{equation}

\begin{figure*}[htb]
\centering
\centerline{\includegraphics[width=17cm]{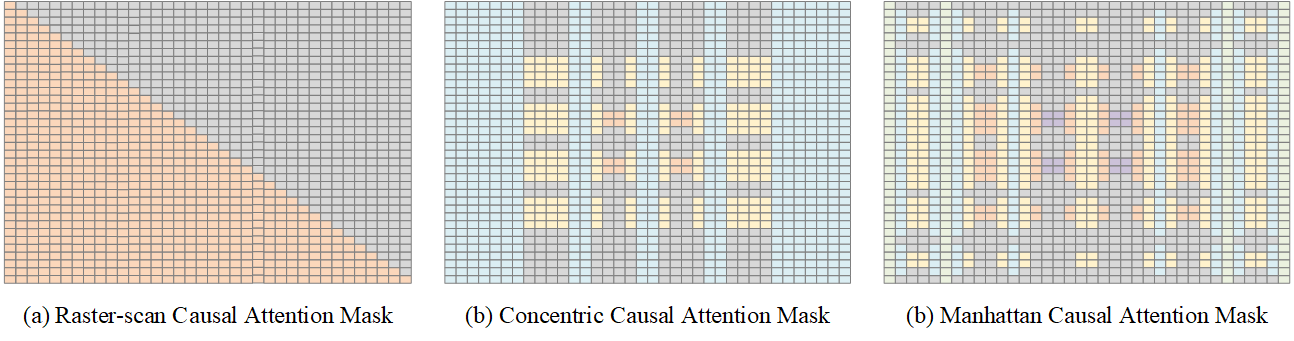}}
\caption{The default causal masks are $V \times V$. We show the causal masks when the number of image tokens $V$ is 36. By default causal modeling in (a), image tokens focus on all visual tokens in between; by CCA in (b), the central image tokens focus on peripheral tokens; and by MCA in (c), image tokens focus on neighboring image tokens in four directions.}
\label{Masking}
\vspace{-4pt}
\end{figure*}

Therefore, we use the sum of token coordinates as new position indexes, termed Manhattan positions assignment $\mu$, to adapt the calculation of the relative positional distance of Manhattan. As shown in Figure \ref{coordinates}.b, Manhattan position assignment preserves the local spatial properties of the image: causal attention of image tokens extends from unidirectional ROPE decay to multidirectional decay. Additionally, compared to raster scanning, the number of Manhattan Position indexes decreases from $V$ to $\sqrt{v}-1$, which reduces the overall distance between image and instruction tokens. This is more favorable for information interaction\cite{cca-llava}.

% \begin{figure}[htb]
% \centering
% \centerline{\includegraphics[width=7cm]{Manhattan Positions Assignment.pdf}}
% \caption{attention mechanism}
% \label{fig:res}
% %
% % \vspace{-15pt}
% \end{figure}

\subsection{Manhattan Causal Masking}

The default causal attention scores  $Attn$ between $Q_i$ and $K_j$ are calculated by Eq.2. We propose MCA to model the relative positions of tokens by 2D Manhattan distance, updating $Attn$ to $Attn^{'}$:
\begin{equation}
\resizebox{\columnwidth}{!}{
$  Attn^{'}_{i,j} = \text{softmax}\left( \frac{Q_i^T \cdot \left( R^d_{\theta,(x_i,y_i)} \right)^T \cdot R^d_{\theta,(x_j,y_j)} \cdot K_j}{\sqrt{d}} \right) = \text{softmax}\left( \frac{Q_i^T \cdot R^d_{\theta,{(x_j - x_i)+(y_j - y_i)}} \cdot K_j}{\sqrt{d}} \right)
$ }
\end{equation} 
By Manhattan position assignment and constant transformation (Eq. 7), $Attn^{'}$ is formally unified with $Attn$. We follow the principle of default causal attention, where the query token $Q_i$ can only attend to the previous key tokens $\{K_j, j <= i\}$ in the sequence during causal attention masking, shown in Figure \ref{Masking}.a. Our Manhattan causal masking is presented in Figure \ref{Masking}.c. For the two-dimensional continuity information contained in images, we preserve spatial localization properties when modeling causal attributes, mitigating object hallucination triggered by image observation bias.

\section{Experiment}

MCA was evaluated on popular hallucination benchmarks, including POPE and CHAIR, as well as general-purpose benchmarks, such as GQA, VQA, MME, SQA, etc. Results show that MCA improves overall visual information perception in LVLMs, rather than overfitting to hallucination-specific datasets. LLaVA-1.5-7B was used as the baseline LVLM, and MCA was further extended to different model architectures (e.g., InternVL-7B) and larger model sizes (e.g., LLaVA-1.5-13B) to verify the robustness. Ablation studies on different positional encoding methods and MCA variants further confirm the effectiveness of the proposed approach.

% \begin{table*}[t]
% \centering
% \scalebox{0.85}{
% \begin{tabular}{lcccccccc}
% \toprule
% Model & VQA$^2$ & GQA & VizWiz & SQA$^\dagger$ & TextVQA  & MMB & MMStar  & MM-Vet \\
% \midrule
% LLaVA1.5-7B & 78.5 & 62.0 & 50.0 & 66.8 & 46.05  & 64.3 & 30.0  & 31.1 \\
% VCD  & 78.6 & 63.5 & 53.7 & 67.3 & -  & - & 34.6  & 33.7 \\
% CCA  & 78.6 & 63.5 & 53.7 & 67.3 & 45.94  & 64 & 33.2  & 33.7 \\
% \rowcolor{gray!20} 
% Ours  & - & 62.97 & 53.54& 68.72 & 45.56  & \textbf{65.81} & \textbf{36.5}  & - \\
% \bottomrule
% \end{tabular}}
% \caption{Performance comparison on various benchmarks using a three-line table.}
% \label{benchmark}
% \end{table*}

\subsection{Experimental Setup and Dataset}
\noindent\textbf{Training Details.}
All experiments are performed on an 8xA800. The visual encoder uses the pre-trained CLIP\cite{radford2021learning} ViT-L/14 and the LLM uses Vicuna-7B\cite{vicuna}. We adopt two-stage training: pre-training stage on CC-558K dataset\cite{llava} with 1 epoch and 256 batch size; instruction tuning stage on 665k multi-turn conversation dataset\cite{llava1.5} with 1 epoch and 128 batch size.
% All experiments are performed on an 8xA800. The visual encoder uses the pre-trained CLIP\cite{radford2021learning} ViT-L/14 with the image resolution of 336x336 and the LLM uses Vicuna-7B\cite{vicuna}. We adopt two-stage training: pre-training stage on CC-558K dataset\cite{llava} with 1 epoch and 256 batch size; instruction tuning stage on 665k multi-turn conversation dataset\cite{llava1.5} with 1 epoch and 128 batch size.

\begin{table}[t]
  \centering
  \scalebox{0.85}{
  \begin{tabular}{c|cc|cccc}
    \toprule
    \rowcolor{gray!20}
     & \multicolumn{2}{c|}{\textbf{POPE}} & \multicolumn{4}{c}{\textbf{CHAIR}} \\
    \rowcolor{gray!20}
    \multirow{-2}{*}{\cellcolor{gray!20}\textbf{Methods}} & \textit{\textbf{F1-score}}$\uparrow$ & \textit{\textbf{acc}}$\uparrow$ & $C_S$ $\downarrow$ & $C_I$ $\downarrow$ & \textbf{Recall}$\uparrow$ & \textit{\textbf{Avg. Len}} \\
    \midrule
    Greedy Search        & 79.3& 79.8& 47.0   & 13.8  & 76.6 & 94.2 \\
    Beam Search    & 84.9 & 86.0 & 51.0   & 15.2  & 75.2 & 102.2  \\
    DoLa  \cite{dola}         & 80.2 & 83.1& 57.0   & 15.2  & 78.2 & 97.5 \\
    ITI \cite{ITI}   & 83.7 & 84.9& 48.2   & 13.9  & 78.3 & 98.6 \\
    VCD   \cite{vcd}         & 83.2 & 82.0& 51.0   & 14.9  & 77.2 & 101.9  \\
    % AGLA \cite{agla} &   84.6 & 43.0   & 14.1 & 78.9 & 98.8 \\
    OPERA \cite{opera}         & 85.2 & 84.2& 47.0   & 14.6  & 78.5 & 95.3 \\
    DOPRA \cite{DOPRA} & 85.6 & 84.3& 46.3   & 13.8  & 78.2 & 96.1 \\
    HALC \cite{halc} & 83.9 & 84.0 & 50.2 & 12.4 & 78.4 & 97.2 \\
    % FastV \cite{fastv} & 81.3 & \underline{39.4} & \underline{11.3} & 69.5 & 90.0 \\
    Less is more \cite{less} & \textbf{86.0} & \textbf{86.8}& \underline{40.2}   & 12.3  & 75.7 & 79.7 \\
    CCA-LLaVA \cite{cca-llava} & \underline{85.9} & \underline{86.5} & 43.0   & \underline{11.5}  & 80.4 & 96.6 \\
     TAME \cite{tame} & 85.5 & 85.9 & 45.2  & 14.0  & 74.4 & 98.8 \\
    SID \cite{sid} & 85.6 & 85.8 & 44.2 &12.2 &73.0 & 99.4 \\
    % CCA-LLaVA(train-free)  & 84.7 & 84.5 & 51.2   & 15.1  & 79.1 & 97 \\
    % See what you \cite{see-what} & 84.9 & 52.4 & 14.5 & 79.1 & 103.0 \\ 
    % EAH\cite{EAH}  & 85.7 & \textbf{36.4} & \textbf{9.9} & 74.9 & 97.7 \\
     % MCA-LLaVA(train-free)  & 85.2 & 85.59 & 41.6  &12.5  &75.7 & 93.1  \\
     % MCA-LLaVA(sft)  & 85.96 & 86.47 & \textbf{38.0}  &\textbf{10.9}  &76.6 & 92.5  \\
    %  \rowcolor{blue!10} 
    % MCA-LLaVA(train-free)  & 85.2 & 85.6 & 41.6  &12.5  &75.7 & 93.1  \\     
    \rowcolor{gray!10} 
   \textbf{ MCA-LLaVA}  & \textbf{86.0} & \underline{86.5} & \textbf{38.0}  &\textbf{10.9}  &76.6 & 92.5  \\
    \bottomrule
  \end{tabular}}
    \caption{Compare results of MCA with other SOTA methods on POPE and CHAIR datasets. We report the average $F1-score$ computed on random, popular, and adversarial splits of POPE (baseline: LLaVA-1.5-7B), max-tokens=512. The best performances within each setting are \textbf{bolded}.}
      \label{Hallucination}
\vspace{-17pt}
\end{table}

\subsection{Evaluation Results of MCA-LLaVA on Hallucination Benchmark}

\begin{table*}[t]
\centering
% \scalebox{0.8}{
\begin{tabular}{lccccccccccc}
\toprule
\rowcolor{gray!20}
Model  & GQA & VizWiz & SQA$^\dagger$ &  MMB & MMStar  & VQA$^{v2}$  &SEED$^{A}$ &SEED$^{I}$ &SEED$^{V}$ \\
\midrule
LLaVA1.5-7B  & 62.0 & 50.0 & 66.8   & 64.3 & 30.0 & 78.5  & 58.6&66.1 &37.3\\
VCD   & 61.9 & 50.5 & 68.5 & -   & 34.6   & 58.3& 63.7 & 37.6\\
CCA-LLaVA   & \textbf{63.5} & \textbf{53.7} & 67.3  & 64.0 & 33.2 & - & 61.7 & 67.1&41.0 \\
 \rowcolor{gray!10}  
% Ours   & 62.97 & 53.54& 68.72 & 45.56  & \textbf{65.81} & \textbf{36.5}  & - \\
\textbf{MCA-LLaVA }  & \textbf{63.0(\textcolor{blue}{+1.0})} & \textbf{53.6(\textcolor{blue}{+3.6})}& \textbf{68.7(\textcolor{blue}{+3.5})}   & \textbf{65.8(\textcolor{blue}{+1.5})} & \textbf{36.5(\textcolor{blue}{+6.5})} & \textbf{78.9(\textcolor{blue}{+0.1})}  &\textbf{62.1(\textcolor{blue}{+3.5})} & \textbf{67.9(\textcolor{blue}{+1.8})}& \textbf{41.3(\textcolor{blue}{+4.0})}\\
\bottomrule
\end{tabular}
\caption{Performance comparison on six general vision-language tasks. These benchmarks include multiple-choice questions from different domains. The experiments were conducted using lmms-eval on A800.}
\label{benchmark}
\vspace{-9pt}
\end{table*}

\noindent\textbf{Evaluation Benchmarks.} 
POPE\cite{pope} evaluates LVLMs hallucination through object-level question-answering tasks. Check whether the model correctly identifies the presence of a specific object in the image by querying prompts like \texttt{"Is there a <object> in the image?"}. CHAIR \cite{chair} evaluates LVLMs hallucination through object-level image captioning tasks. It includes two evaluation aspects: instance-level hallucinations CHAIR$_{I}$ ($C_I$) and sentence-level hallucinations CHAIR$_{S}$ ($C_S$), calculated as follows:
\begin{equation}
    \scriptsize C_S = \frac{|\{\text{hallucinated objects}\}|}{|\{\text{all mentioned objects}\}|}
\end{equation}
\begin{equation}
     \scriptsize C_I = \frac{|\{\text{captions w/ hallucinated objects}\}|}{|\{\text{all captions}\}|}
\end{equation}

% \begin{figure}[t]
% % \vspace{-10pt}
% \centering
% \centerline{\includegraphics[width=8cm]{images/radar.pdf}}
% \caption{Experimental results on POPE~\cite{pope}, CHAIR~\cite{chair}, and MME~\cite{mme} demonstrate the excellent performance of MCA-LLaVA in mitigating hallucinations.}
% \label{fig:res}
% % \vspace{-10pt}
% \end{figure}

\begin{table}[h]
\centering
\footnotesize
\scalebox{0.95}{
\begin{tabular}{@{}cccccc}
\toprule
\rowcolor{gray!20}
 & \multicolumn{2}{c}{\cellcolor{gray!20}\textbf{Object-level}} & \multicolumn{2}{c}{\cellcolor{gray!20}\textbf{Attribute-level}} &  \\
\rowcolor{gray!20}
\multirow{-2}{*}{\cellcolor{gray!20}\textbf{Methods}}& \multicolumn{1}{c}{\cellcolor{gray!20}\textit{\textbf{Existence}}$\uparrow$} & \multicolumn{1}{c}{\cellcolor{gray!20}\textit{\textbf{Count}}$\uparrow$} & \multicolumn{1}{c}{\cellcolor{gray!20}\textit{\textbf{Position}}$\uparrow$} & \multicolumn{1}{c}{\cellcolor{gray!20}\textit{\textbf{Color}}$\uparrow$} & \multirow{-2}{*}{\cellcolor{gray!20}\textbf{Total Score}$\uparrow$} \\ \midrule
 Beam                    & 175.67 & 124.67 & 114.00 & 151.00 & 565.34 \\
 Greedy                    & 185.00 & 93.33 & 110.00 & 156.67 & 545.00 \\
 DOLA  \cite{dola}                  & 175.00 & 108.33 & 90.00 & 138.33 & 511.66 \\
 VCD   \cite{vcd}                    & 184.66 & 138.33& 128.67 & 153.00 & 604.66 \\
 OPERA   \cite{opera}                  & 180.67 & 133.33 & 123.33 & 155.00 & 592.33 \\
CCA-LLaVA\cite{cca-llava}            & 190.00 & 148.33 & 128.33 &\textbf{ 175.00 }& 641.66 \\
SID \cite{sid} & 182.00 & 127.00 & 116.00 & 139.00 & 564.00\\
TAME+OPERA \cite{tame}        & 176.00 &118.33 & 113.00 & 143.00 &550.33   \\
% EAH  \cite{eah}           & 190.00 & 108.33 & \textbf{145.00} & 160.66 & 603.99 \\ 
 \rowcolor{gray!10}  
\textbf{MCA-LLaVA (Our)  }                  & \textbf{190.00} & \textbf{163.33} & 126.67 & 170.00 & \textbf{650.00}  \\
\bottomrule
\end{tabular}
}
\caption{Evaluation results on the hallucination subset of MME \cite{mme}. 
% Regular decoding denotes direct sampling, whereas VCD refers to sampling from our proposed contrastive distribution pvcdp_{vcd}. 
The best performances within each setting are \textbf{bolded}, baseline: LLaVA1.5-7B.}
\label{mme}
\vspace{-15pt}
\end{table}

\noindent\textbf{Effectiveness of MCA-LLaVA.}
In this paper, the POPE and CHAIR scores of LLaVA1.5-7b under greedy search are used as the baseline results. According to Section 4.2, we replaced the image position index determined by the RoPE raster-scan with the Manhattan positions assignment. This approach introduces spatial priors to RoPE long-term decay, and achieves a 6.7\% improvement in F1 score and a 6.7\% improvement in accuracy on POPE, as shown in Table \ref{Hallucination}. We hypothesize that MCA-LLaVA alleviates image observation bias, allowing the instruction tokens to focus more effectively on the image information. Compared with other state-of-the-art hallucination mitigation methods, the MCA-LLaVA obtained the highest F1 score of 86.0\% and the second highest accuracy of 86.5\% on POPE.

% \begin{table*}[t]
% \centering
% % \scalebox{0.8}{
% \begin{tabular}{lccccccccccc}
% \toprule
% \rowcolor{gray!20}
% Model  & GQA & VizWiz & SQA$^\dagger$ &  MMB & MMStar  & VQA$^{v2}$ & TextVQA &SEED$^{A}$ &SEED$^{I}$ &SEED$^{V}$ \\
% \midrule
% LLaVA1.5-7B  & 62.0 & 50.0 & 66.8   & 64.3 & 30.0 & 78.5 & \textbf{58.2} & 58.6&66.1 &37.3\\
% VCD   & 61.9 & 50.5 & 68.5 & -   & 34.6  & - & 54.4 & 58.3& 63.7 & 37.6\\
% CCA-LLaVA   & \textbf{63.5} & \textbf{53.7} & 67.3  & 64.0 & 33.2 & - & 57.8& 61.7 & 67.1&41.0 \\
%  \rowcolor{gray!10}  
% % Ours   & 62.97 & 53.54& 68.72 & 45.56  & \textbf{65.81} & \textbf{36.5}  & - \\
% \textbf{MCA-LLaVA }  & 63.0(\textcolor{blue}{+1.0}) & 53.6& \textbf{68.7(\textcolor{blue}{+3.5})}   & \textbf{65.8(\textcolor{blue}{+1.5})} & \textbf{36.5(\textcolor{blue}{+6.5})} & \textbf{78.9(\textcolor{blue}{+0.1})} & 58.1 &\textbf{62.1} & \textbf{67.9}& \textbf{41.3}\\
% \bottomrule
% \end{tabular}
% \caption{Performance comparison on six general vision-language tasks. These benchmarks include multiple-choice questions from different domains. The experiments were conducted using lmms-eval on A800.}
% \label{benchmark}
% % \vspace{-9pt}
% \end{table*}

For CHAIR, we set the maximum number of generated response tokens to 512 to evaluate the generated long captions hallucinations. It is important to note that longer responses better reflect the model's perception of the image information. As shown in Table \ref{Hallucination}, MCA-LLaVA achieves the best $C_I$ and $C_S$ among all SOTA methods. Under greedy search, our model improves 9\% at the sentence level and 2.9\% at the instance level compared to the baseline. These results demonstrate the importance of optimizing RoPE image tokens position modeling for mitigating hallucinations.

\subsection{Evaluation Results of MCA-LLaVA on General Vision-language Benchmarks}

\noindent\textbf{Evaluation Benchmarks.} We also evaluate MCA-LLaVA on more visual-language benchmarks, including general visual-linguistic tasks and vision-centered tasks such as MME-Bench \cite{mme}, VizWiz \cite{vizwiz}, MMSTAR \cite{mmsatr}, GQA \cite{gqa}, SEED\cite{Li2023SEEDBenchBM}, TextVQA\cite{textvqa}, VQAv2\cite{Goyal2016MakingTV}, and ScienceQA \cite{sqa}. 

% MME-Bench also examines the general perception capabilities of LVLMs using a wide range of tasks. Additionally, we compare against VizWiz and GQA, which assess specific perception capabilities, such as knowledge and relations.
% TextVQA \cite{textvqa}
% \begin{table}[t]
% \centering
% \scalebox{0.65}{
% \begin{tabular}{lcccccc}
% \toprule
% Model  & GQA & VizWiz & SQA$^\dagger$ & TextVQA  & MMB & MMStar   \\
% \midrule
% LLaVA1.5-7B  & 62.0 & 50.0 & 66.8 & 46.05  & 64.3 & 30.0   \\
% VCD   & 61.9 & 50.5 & 68.5 & -  & - & 34.6   \\
% CCA-LLaVA   & 63.5 & 53.7 & 67.3 & 45.94  & 64.0 & 33.2   \\
% \rowcolor{gray!20} 
% % Ours   & 62.97 & 53.54& 68.72 & 45.56  & \textbf{65.81} & \textbf{36.5}  & - \\
% MCA-LLaVA   & 63.0 & 53.6& \textbf{68.7} & 45.56  & \textbf{65.8} & \textbf{36.5}  \\
% \bottomrule
% \end{tabular}}
% \caption{Performance comparison on six general vision-language tasks. These benchmarks include multiple-choice questions from different domains. The experiments were conducted using lmms-eval on A800.}
% \label{benchmark}
% \end{table}

\begin{figure}[t]
\vspace{-15pt}
\centering
\centerline{\includegraphics[width=8cm]{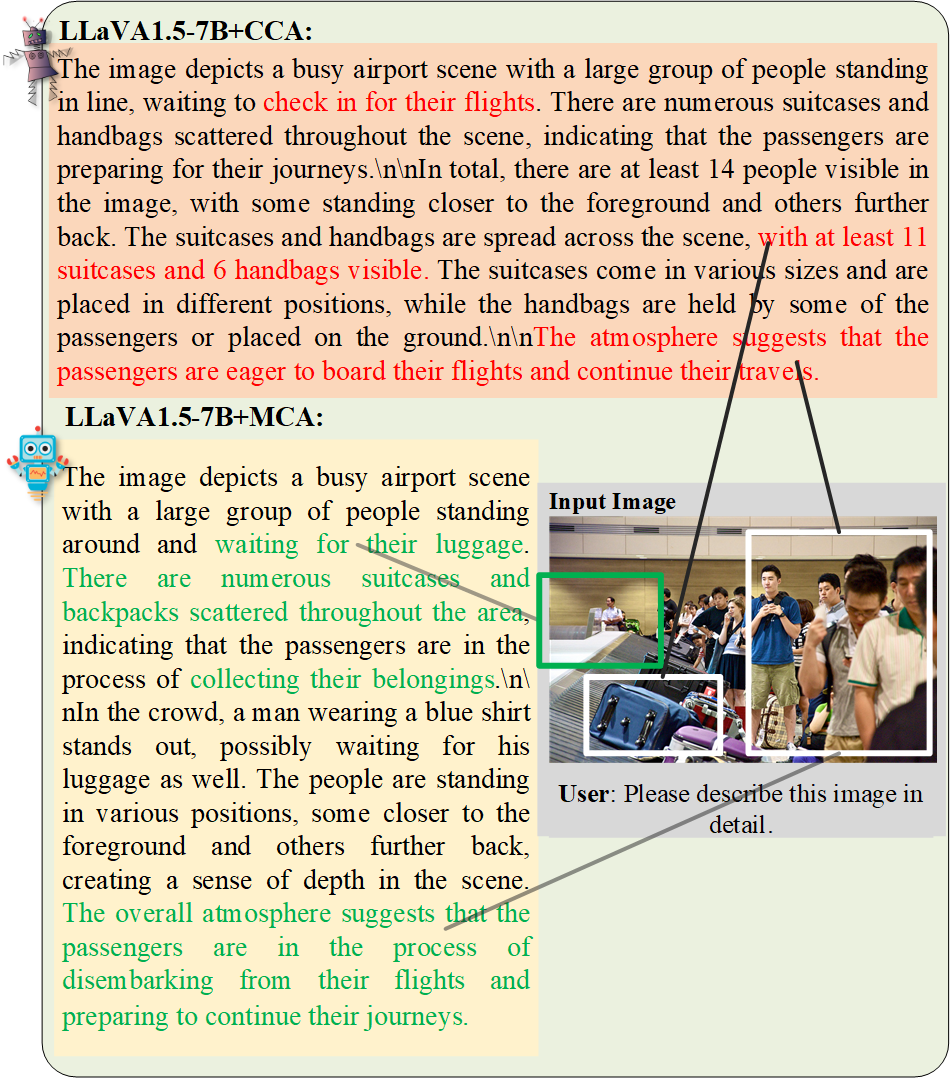}}
\caption{Qualitative results of CCA-LLaVA with MCA-LLaVA. Our method generate less hallucinations.}
\label{sample}
\vspace{-9pt}
\end{figure}

\noindent\textbf{Effectiveness of MCA-LLaVA.}
MME evaluates the overall image perception ability of the model, revealing hallucinations from a broader perspective, not just object hallucinations. We analyze the object-level and attribute-level hallucinations of LVLMs from four types of metrics: object existence, count, position, and color. We report the results of MCA-LLaVA on MME in Table \ref{mme}. MCA-LLaVA outperforms the baseline model in all four evaluation metrics, with a total score improvement of 105 compared to the baseline. Additionally, MCA-LLaVA showed better image perception and total score compared to other SOTA. Notably, our method outperforms CCA-LLaVA by 15 on the Count metric, indicating that our approach can perceive the overall image information at a finer granularity. The qualitative examples in Figure \ref{sample} reveal the hallucinations generated by CCA-LLaVA in analyzing object counts in graphical scenes.

To evaluate the model's general perception ability beyond hallucinations, we tested the performance of MCA-LLaVA on eight general vision-language tasks using lmms-eval. These benchmarks evaluate the model through multiple-choice questions, covering topics such as scientific knowledge, complex reasoning, and more. As shown in table \ref{benchmark}, MCA-LLaVA demonstrates consistent performance improvements across all benchmarks, such as a 1.5\% and 6.5\% increase over the baseline on MMB and MMStar, respectively. The experimental results demonstrate that MCA-LLaVA enhances the model's image perception ability comprehensively by mitigating image bias, rather than overfitting to hallucination benchmarks.

\begin{figure*}[t]
\centering
\centerline{\includegraphics[width=17.5cm]{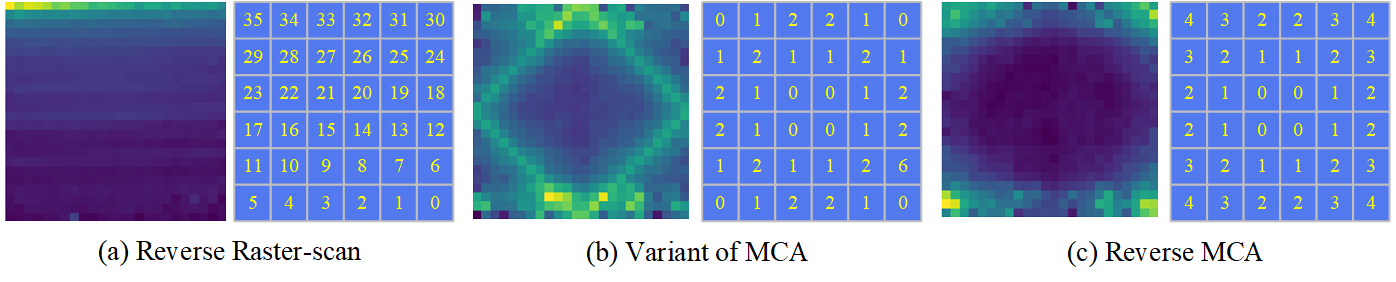}}
\caption{Different positional indices and corresponding information flow patterns.}
\label{position}
%
% \vspace{-9pt}
\end{figure*}

\subsection{Results of MCA-LLaVA with CCA-LLaVA}

As shown in Figure \ref{sample}, the description generated by CCA-LLaVA contains hallucinated elements. It focuses on the people queuing at airports but incorrectly describes the process of preparing passengers to board a plane. Additionally, the generated response contains two incorrect descriptions of item quantities: 11 suitcases and 6 handbags. This reveals the model's insufficient understanding of the global image information.

On the other hand, MCA-LLaVA's description correctly states the fact that people are waiting to collect their luggage. This suggests that MCA-LLaVA better integrates the overall image information and understands the context of disembarking from the plane. Additionally, MCA-LLaVA's description mentions objects that CCA-LLaVA does not, such as a man in a blue shirt and backpack. This indicates that MCA-LLaVA is capable of observing richer and more fine-grained image information.

\subsection{Ablation Study}

\begin{table}
  \centering
   \scalebox{0.81}{
  \begin{tabular}{c|cc|ccc}
    \toprule
    \rowcolor{gray!20}  
     & \multicolumn{2}{c|}{\textbf{POPE} }  & \multicolumn{3}{c}{\textbf{CHAIR }} \\
    \rowcolor{gray!20}  
  \multirow{-2}{*}{\textbf{Method}}  & \textbf{F1 score}$\uparrow$  & \textbf{acc}$\uparrow$ &\textbf{ \textit{C\( _{S} \)}} $\downarrow$ &\textbf{\textit{ C\( _{I} \)}} $\downarrow$ & \textit{\textbf{Recall}}$\uparrow$  \\
    \midrule
    LLaVA-1.5-7B        & 79.3& 79.8& 47.0   & 13.8 &76.6  \\
     \rowcolor{gray!10}
   \textbf{MCA} & 86.0 (\textcolor{blue}{+6.7})& 86.5 (\textcolor{blue}{+6.7})& 38.0(\textcolor{blue}{+9.0})  & 10.9(\textcolor{blue}{+2.9})  & 76.6 \\
    LLaVA-1.5-13B    & 82.4 &  82.7 & 44.0    &  12.7  & 77.3  \\
     \rowcolor{gray!10}
    \textbf{MCA}    & 85.9(\textcolor{blue}{+3.5}) &  86.1(\textcolor{blue}{+3.4}) &  37.2(\textcolor{blue}{+6.8})   & 10.3(\textcolor{blue}{+2.4}) & 78.0 \\
    InternVL-7B  & 81.6 & 82.2 & 45.8   & 12.9 &79.1 \\
     \rowcolor{gray!10}
    \textbf{MCA}  & 83.1(\textcolor{blue}{+1.5}) & 83.6(\textcolor{blue}{+1.4}) & 44.9(\textcolor{blue}{+0.9})   & 12.3(\textcolor{blue}{+0.6}) & 80.3 \\
    \bottomrule
  \end{tabular}}
    \caption{Generalization study of MCA on other LVLMs.}
      \label{Generalization}
\vspace{-9pt}
\end{table}

% \begin{table}
%   \centering
%    \scalebox{0.78}{
%   \begin{tabular}{c|cc|cc}
%     \toprule
%     \multirow{2}{*}{Model} & \multicolumn{2}{c|}{POPE }  & \multicolumn{2}{c}{CHAIR } \\
   
%     & F1 score$\uparrow$  & acc$\uparrow$ & \textit{C\( _{S} \)} $\downarrow$ &\textit{ C\( _{I} \)} $\downarrow$  \\
%     \midrule
%     LLaVA-1.5-7B        & 79.3& 79.8& 47.0   & 13.8  \\
%     LLaVA-1.5-7B+\textbf{MCA} & 86.0& 86.5& 38.0  & 10.9  \\
%     LLaVA-1.5-13B    & 82.4 &  82.7 & 44.0    &  12.7   \\
%     LLaVA-1.5-13B+\textbf{MCA}    & 85.9 &  86.1 &  37.2   & 10.3   \\
%     InternVL-7B  & 81.6 & 82.2 & 45.8   & 12.9  \\
%     InternVL-7B+\textbf{MCA}  & 83.1 & 83.6 & 44.9   & 12.3  \\
%     \bottomrule
%   \end{tabular}}
%     \caption{Generalization study of MCA on other LVLMs models.}
%       \label{Generalization}
% \vspace{-9pt}
% \end{table}

\textbf{Generalization Study of MCA}
To further validate the robustness of the proposed method, MCA was applied to additional LVLMs. Similar to the baseline model LLaVA-1.5-7B, InternVL-7B also adopts a RoPE-based positional modeling mechanism. InternVL leverages LLaMA2 to construct QLLaMA, enabling more effective alignment between visual and language modalities. Although the original InternVL-7B already demonstrates strong performance, the integration of MCA further enhances its performance. Additionally, experiments were conducted on the larger LLaVA-1.5-13B model. As shown in the table, MCA can obtain performance enhancement under different scale models, proving its robustness. Results indicate that MCA continues to mitigate hallucinations as model scale increases, further demonstrating its robustness.

\noindent\textbf{Ablation Study of Position Index Settings}
As shown in Figure \ref{position} and Table \ref{PositionAblation}, MCA-LLaVA preserves two-dimensional local spatial features by assigning the origin of position coordinates to the tokens at the four corners of the image and computing the Manhattan relative distance between tokens. We conduct two sets of ablation studies to verify the effectiveness of this coordinate assignment strategy: (1) setting both the image center and the four corners as origins; (2) setting only the image center as the origin. The results in Table \ref{PositionAblation} confirm that the coordinate design in MCA is the optimal configuration. In addition, compared with the CCA and Reverse Raster-scan settings, our method improves position modeling and addresses the long-range decay of RoPE through a relative position computation mechanism, rather than relying on heuristic position reassignments, making it more effective and interpretable.

\begin{table}
  \centering
   \scalebox{0.73}{
  \begin{tabular}{c|c|cc|ccc}
    \toprule
    \rowcolor{gray!20}  
   &  & \multicolumn{2}{c|}{\textbf{POPE} }  & \multicolumn{3}{c}{\textbf{CHAIR} } \\
   \rowcolor{gray!20}  
    \multirow{-2}{*}{\textbf{Method}}&\multirow{-2}{*}{\textbf{Num}} &\textbf{F1 score}$\uparrow$  & \textbf{acc}$\uparrow$ & \textbf{\textit{C\( _{S} \)}} $\downarrow$ &\textbf{\textit{ C\( _{I} \)}} $\downarrow$ & \textit{\textbf{Recall}}$\uparrow$  \\
    \midrule
    Raster-scan      &526 & 79.3& 79.8& 47.0   & 13.8  &76.6 \\
    Reverse Raster-scan   & 526  & 76.1 & 76.6  &  48.1   &  14.1 & 75.2 \\
    CCA  & 12 & 85.9 & 86.5 & 43.0   & 11.5 & 80.4\\
    Variant of MCA  & 12& 81.3  & 81.2  & 52.4   & 14.2  & 81.9\\   
    Reverse MCA  &23 & 80.8  & 81.2  & 50.8   & 14.4 & 74.9 \\   
    \rowcolor{gray!10} 
   \textbf{ MCA}   &23& 86.0(\textcolor{blue}{+6.7}) & 86.5(\textcolor{blue}{+6.7}) & 38.0(\textcolor{blue}{+9.0})  & 10.9(\textcolor{blue}{+2.9})   &76.6\\
    \bottomrule
  \end{tabular}}
    \caption{Ablation experiments under different positional coding methods and MCA variants. Num denotes the number of image tokens positional indexes.}
    % \caption{Ablation experiments under different positional coding methods and MCA variants. Num denotes the number of image tokens positional indexes under different positional encodings}
      \label{PositionAblation}
\vspace{-9pt}
\end{table}

\begin{figure}[t]
\vspace{-7pt}
\centering
\centerline{\includegraphics[width=8.5cm]{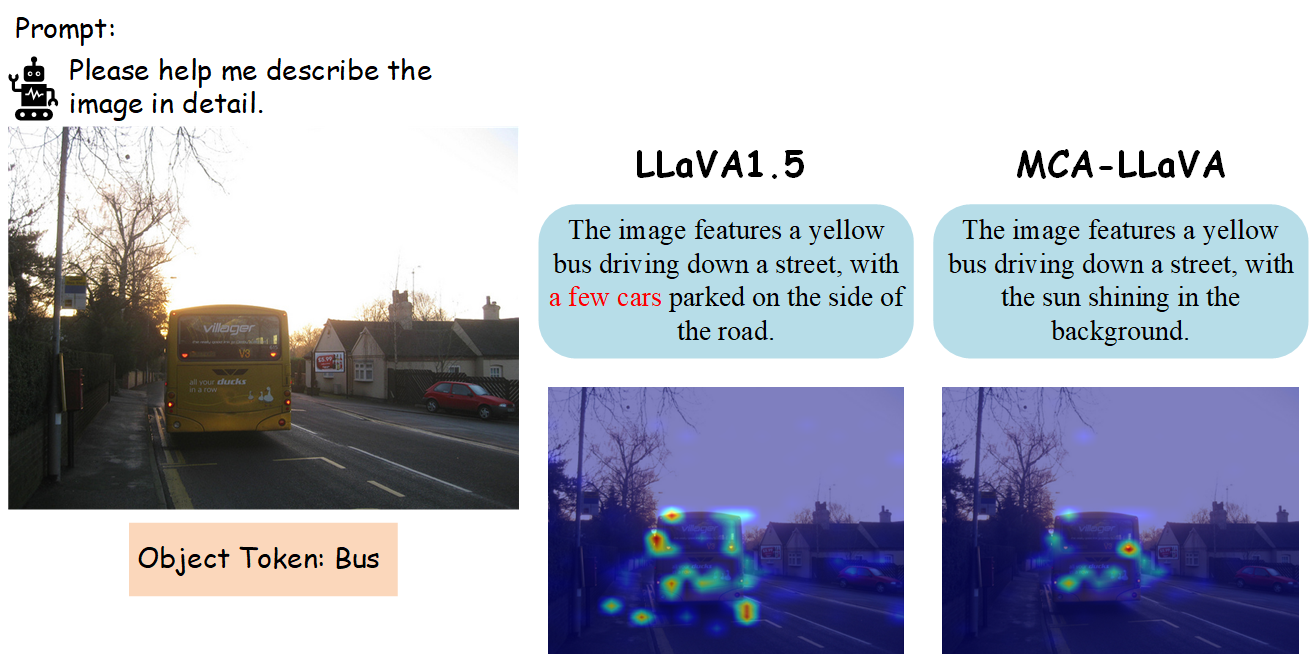}}
\caption{Attention Maps Visualization of MCA.}
\label{cam}
\vspace{-13pt}
\end{figure}

\noindent\textbf{Attention Maps Visualization of MCA}
We further analyze heatmaps over the image for object tokens to investigate the model's differences in visual perception. As shown in the Figure \ref{cam}, compared to the baseline, MCA-LLaVA demonstrates increased attention to local features of objects and enhanced perception of other regions in the image. We hypothesize that MCA helps improve the model’s visual perception capabilities, including both global context and local details, thereby contributing to the mitigation of hallucinations.

\section{Conclusion}

This paper provides an in-depth analysis of the limitations of the one-dimensional long-term decay of RoPE: the long-term decay induces image alignment bias, where image tokens distant from the instruction tokens are considered unimportant. With the help of information flow, we find that image alignment bias hinders cross-modal alignment and makes LVLMs more prone to hallucinations. To this end, we improve the RoPE relative position calculation mechanism and propose the  Manhattan Causal Attention (MCA). The results of multiple evaluation benchmarks demonstrate the effectiveness of MCA.

% \section{Limations}

% Due to computational resource limitations, we did not evaluate the MCA on larger LVLMs. Additionally, our approach extends the one-dimensional long-term decay to the spatial domain, future discussions will be made for more modalities (e.g., audio, video).

% \section{Acknowledgments}

% Identification of funding sources and other support, and thanks to
% individuals and groups that assisted in the research and the
% preparation of the work should be included in an acknowledgment
% section, which is placed just before the reference section in your
% document.

% \begin{acks}
% \end{acks}

\bibliographystyle{ACM-Reference-Format}
\bibliography{sample-base}

% \appendix



\end{document}